\documentclass[11pt,a4paper]{article}
\usepackage{times,latexsym}
\usepackage{times}
\usepackage{latexsym}
\usepackage{framed}
\usepackage{amsmath} 
\usepackage{amssymb} 
\usepackage{graphicx}
\usepackage{dsfont}
\usepackage{bm}
\usepackage{amsfonts}
\usepackage{listings}
\usepackage{caption}
\usepackage{subcaption}
\usepackage{pythonhighlight}
\usepackage{booktabs}
\usepackage{pifont}
\usepackage{url}
\usepackage[T1]{fontenc}
\usepackage{multirow}
\usepackage{tabularx}
\usepackage{siunitx}

\newcommand{\summac}{\textsc{SummaC}} 

\newcommand{\summacconv}{\summac{}\textsubscript{Conv}}
\newcommand{\summaczs}{\summac{}\textsubscript{ZS}}

\usepackage[acceptedWithA]{tacl2018v2}

\usepackage{xspace,mfirstuc,tabulary}

\newif\iftaclinstructions
\taclinstructionsfalse 
\iftaclinstructions
\renewcommand{\confidential}{}
\newcommand{\instr}
\fi

\iftaclpubformat 

\else

\fi

\title{\summac: Re-Visiting NLI-based Models for\\
Inconsistency Detection in Summarization}

\author{Philippe Laban \\
  UC Berkeley \\
   \\
  \And
  Tobias Schnabel \\
  Microsoft\\
  \And
  Paul N. Bennett\\
  Microsoft
  \And
  Marti A. Hearst \\
  UC Berkeley\thanks{~Author emails: \{phillab,hearst\}@berkeley.edu, \{Tobias.Schnabel,Paul.N.Bennett\}@microsoft.com} \\
}

\date{}

\begin{document}
\maketitle
\begin{abstract}
In the summarization domain, a key requirement for summaries is to be factually consistent with the input document. Previous  work has found that natural language inference (NLI) models do not perform competitively when applied to inconsistency detection. In this work, we revisit the use of NLI for inconsistency detection, finding that past work suffered from a mismatch in input granularity between NLI datasets (sentence-level), and  inconsistency detection (document level). We provide a highly effective and light-weight method called \summacconv{} that enables NLI models to be successfully used for this task by segmenting documents into sentence units and aggregating scores between pairs of sentences. On our newly introduced benchmark called \textbf{\summac{}} (\textbf{Summa}ry \textbf{C}onsistency) consisting of six large inconsistency detection datasets, \summacconv{} obtains state-of-the-art results with a balanced accuracy of 74.4\%, a 5\% point improvement compared to prior work.
\end{abstract}

\section{Introduction}

\begin{figure}
    \centering
    \includegraphics[width=0.45\textwidth]{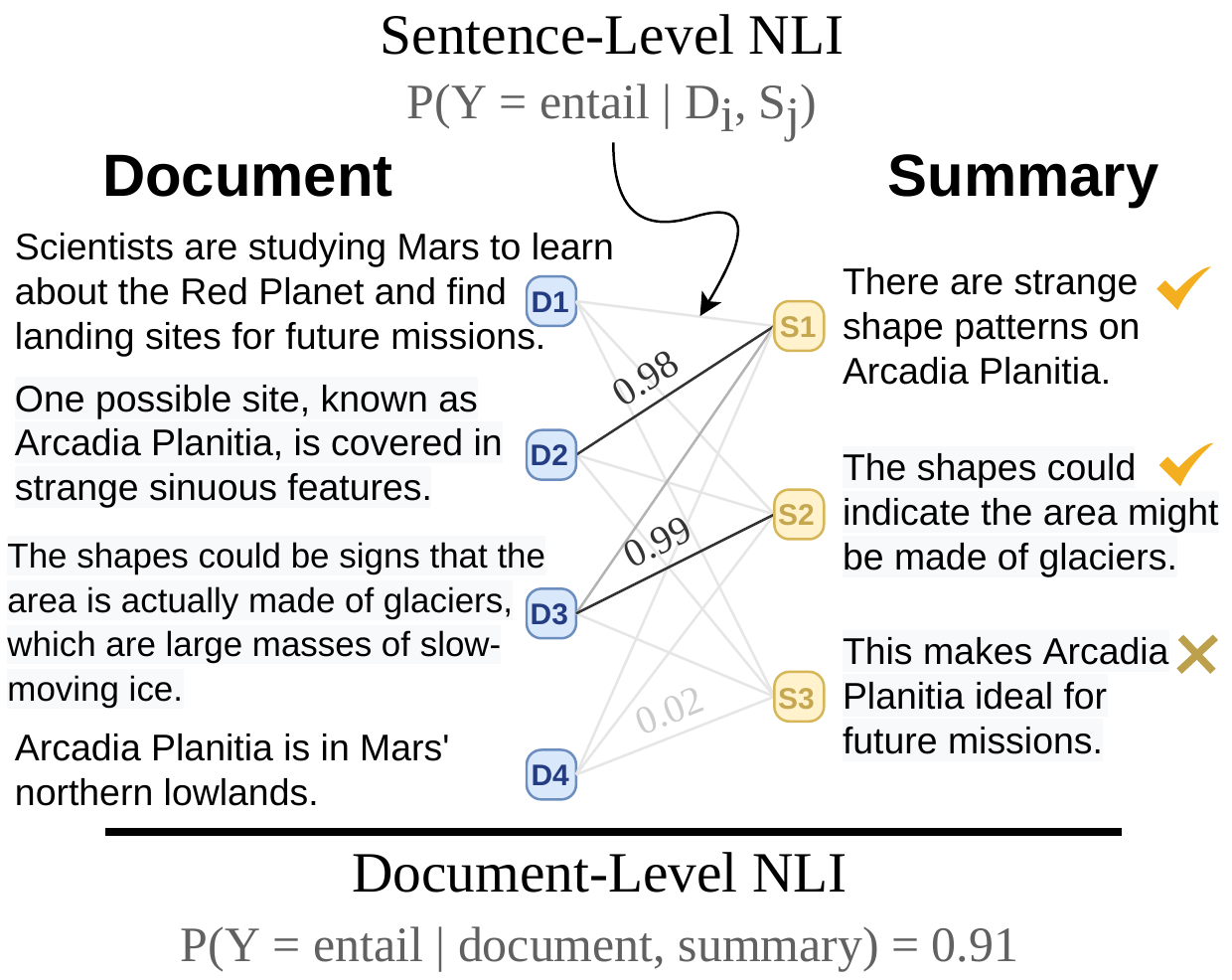}
    \caption{\textbf{Example document with an inconsistent summary.} When running each sentence pair ($D_i, S_j$) through an NLI model, $S_3$ is not entailed by any document sentence. However, when running the entire (document, summary) at once, the NLI model incorrectly predicts that the document highly entails the entire summary.}
    \label{fig:intro_example}
\end{figure}

Recent progress in text summarization has been remarkable, with ROUGE record-setting models published every few months, and human evaluations indicating that automatically generated summaries are matching human-written summaries in terms of fluency and informativeness \cite{zhang2020pegasus}.

A major limitation of current summarization models is their inability to remain factually consistent with the respective input document. Summary inconsistencies are diverse -- from inversions (i.e., negation) to incorrect use of an entity (i.e., subject, object swapping), or hallucinations (i.e., introduction of entity not in the original document). Recent studies have shown that in some scenarios, even state-of-the-art pre-trained language models can generate inconsistent summaries in more than 70\% of all cases \cite{Pagnoni2021UnderstandingFI}. This has led to accelerated research around summary inconsistency detection. 

A closely-related task to inconsistency detection is textual entailment, also referred to as Natural Language Inference (NLI), in which a \textit{hypothesis} sentence must be classified as either entailed by, neutral or contradicting a \textit{premise} sentence. Enabled by the crowd-sourcing of large NLI datasets such as SNLI \cite{bowman2015large} and MNLI \cite{williams2018broad}, modern architectures have achieved close to human performance at the task.

The similarity of NLI to inconsistency detection, as well as the availability of high-performing NLI models, led to early attempts at using NLI to detect consistency errors in summaries. These early attempts were unsuccessful, finding that re-ranking summaries according to an NLI model can lead to an increase in consistency errors \cite{falke2019ranking}, or that out-of-the-box NLI models obtain 52\% accuracy at the binary classification task of inconsistency detection, only slightly above random guessing \cite{kryscinski2020evaluating}.

In this work, we revisit this approach, showing that NLI models \emph{can} in fact successfully be used for inconsistency detection, as long as they are used at the appropriate \textit{granularity}. Figure~\ref{fig:intro_example} shows how crucial using the correct granularity as input to NLI models is. An inconsistency checker should flag the last sentence in the summary (shown right) as problematic. When treating the entire document as the premise and the summary as the hypothesis, a competitive NLI model predicts with probability of $0.91$ that the summary is entailed by the document. However, when splitting the documents into sentence premise-hypothesis pairs (visualized as edges in Figure~\ref{fig:intro_example}) the NLI model correctly determines that S\textsubscript{3} is not supported by any document sentence. This illustrates that working with sentence pairs is crucial for making NLI models work for inconsistency detection.

Our contributions are two-fold. First, we introduce a new approach for inconsistency detection based on the aggregation of sentence-level entailment scores for each pair of input document and summary sentences. We present two model variants that differ in the way they aggregate sentence-level scores into a single score. \summaczs{} performs zero-shot aggregation by combining sentence-level scores using \texttt{max} and \texttt{mean} operators. \summacconv{} is a trained model consisting of a single learned convolution layer compiling the distribution of entailment scores of all document sentences into a single score.

Second, to evaluate our approach, we introduce the \summac{} Benchmark by standardizing existing datasets. Because the benchmark contains the six largest summary consistency datasets, it is more comprehensive and includes a broader range of inconsistency errors than prior work. 

The \summac{} models outperform existing inconsistency detection models on the benchmark, with the \summacconv{} obtaining an overall balanced accuracy of 74.4\%, 5\% above prior work. We publicly release the models and datasets\footnote{\url{https://github.com/tingofurro/summac/}}.

\section{Related Work}
We briefly survey existing methods and datasets for fact checking, inconsistency detection, and inconsistency correction.

\subsection{Fact Checking and Verification}
Fact checking is a related task in which a model receives an input claim along with a corpus of ground truth information. The model must then retrieve relevant evidence and decide whether the claim is supported, refuted or if there is not enough information in the corpus \cite{thorne2018fever}. The major difference to our task lies in the different semantics of consistency and accuracy. If a summary adds novel and accurate information not present in the original document (e.g., adding background information), the summary is accurate but inconsistent. In the summary inconsistency detection domain, the focus is on detecting any inconsistency, regardless of its accuracy, as prior work has shown that current automatic summarizers are predominantly inaccurate when inconsistent \cite{maynez2020faithfulness}.

\subsection{Datasets for Inconsistency Detection}
Several datasets have been annotated to evaluate model performance in inconsistency detection, typically comprising up to two thousand annotated summaries. Datasets are most commonly crowd-annotated with three judgements each, despite some work showing that as many as eight annotators are required to achieve high inter-annotator agreement \cite{falke2019ranking}.

Reading the entire original document being summarized is time-consuming, and to amortize this cost, consistency datasets often contain multiple summaries, generated by different models, for the same original document.

Some datasets consist of an overall consistency label for a summary (e.g., FactCC \cite{kryscinski2020evaluating}), while others propose a finer-grained typology with up to 8 types of consistency errors \cite{huang2020have}.

We include the six largest summary consistency datasets in the \summac{} Benchmark, and describe them more in detail in Section~\ref{section:summac_benchmark}.

\subsection{Methods for Inconsistency Detection}

Due to data limitations, most inconsistency detection methods adapt NLP pipelines from other tasks including QAG models, synthetic classifiers, and parsing-based methods.

\textbf{QAG} methods follow three steps: (1) question generation (QG), (2) question answering (QA) with the document and the summary, (3) matching document and summary answers. A summary is considered consistent if few or no questions have differing answer with the document. A key design choice for these methods lies in the source for question generation. \citet{durmus2020feqa} generate questions using the summary as a source, making their FEQA method precision-oriented. \citet{scialom2019answers} generate questions with the document as a source, creating a recall-focused measure. \citet{scialom2021questeval} unite both in QuestEval, by generating two sets of questions, sourced from the summary and document respectively. We include FEQA and QuestEval in our benchmark results.

\textbf{Synthetic classifiers} rely on large, synthetic datasets of summaries with inconsistencies, and use those to train a classifier with the expectation that the model generalizes to non-synthetic summaries. To generate a synthetic dataset, \citet{kryscinski2020evaluating} propose a set of semantically invariant (e.g., paraphrasing) and variant (e.g., sentence negation) text transformations that they apply to a large summarization dataset. \texttt{FactCC-CLS}, the classifier obtained when training on the synthetic dataset, is included in our benchmark results for comparison.

\textbf{Parsing}-based methods generate relations through parsing and compute the fraction of summary relations that are compatible with document relations as a precision measure of summary factuality. \citet{goodrich2019assessing} extract \texttt{(subject, relation, object)} tuples most commonly using OpenIE \cite{etzioni2008open}. In the recent DAE model, \citet{goyal2020evaluating} propose to use arc labels from a dependency parser instead of relation triplet. We include the DAE model in our benchmark results.

\subsection{Methods for Consistency Correction}

Complementary to inconsistency detection, some work focused on the task of mitigating inconsistency errors during summarization. Approaches fall in two categories: Reinforcement Learning (RL) methods to improve models and stand-alone re-writing methods.

\textbf{RL-methods} often rely on an out-of-the-box inconsistency detection model and use reinforcement learning to optimize a reward with a consistency component. \citet{arumae2019guiding} optimize a QA-based consistency reward, and \citet{nan2021improving} streamline a QAG reward by combining the QG and QA model, making it more efficient for RL training. \citet{pasunuru2018multi} leverage an NLI-based component as part of an overall ROUGE-based reward, and \citet{zhang2020optimizing} use a parsing-based measure in the domain of medical report summarization.

\textbf{Re-writing methods} typically operate as a modular component that is applied after an existing summarization model. \citet{cao2020factual} use a synthetic dataset of rule-corrupted summaries to train a post-corrector model, but find that this model does not transfer well to real summarizer errors. \citet{dong2020multi} propose to use a QAG model to find erroneous spans, which are then corrected using a post-processing model.

Since all methods discussed above for consistency correction rely on a model to detect inconsistencies, they will naturally benefit from more accurate inconsistency detectors.

\section{\summac{} Models}
\label{section:summac_models}

We now introduce our \summac{} models for inconsistency detection. The first step common to all models is to apply an out-of-the-box NLI model to generate an \textit{NLI Pair Matrix} for a \texttt{(document, summary)} pair. The two models we present then differ in the way they process this pair matrix to produce a single consistency score for a given summary. We also describe the \summac{} evaluation benchmark, a set of inconsistency detection datasets in Section~\ref{section:summac_benchmark}.  In Section~\ref{section:results}, we measure the performance of the \summac{} models on this benchmark and investigate components of the models, including which NLI model achieves highest performance, which NLI categories should be used, and what textual granularity is most effective.

\subsection{Generating the NLI Pair Matrix}

NLI datasets are predominantly represented at the sentence level. In our pilot experiments, we found that this causes the resulting NLI models to fail in assessing consistency for documents with 50 sentences and more.

This motivates the following approach. We generate an NLI Pair Matrix by splitting a \texttt{(document, summary)} pair into sentence blocks. The document is split into $M$ blocks, each considered a premise labeled from $D_1, ..., D_M$, and the summary is split into $N$ blocks, each considered a hypothesis labeled from $S_1, ..., S_N$.

Each $D_i, S_j$ combination is run through the NLI model, which produces a probability distribution over the three NLI categories $(E_{ij}, C_{ij}, N_{ij})$ for entailment, contradiction and neutral, respectively. If not specified otherwise, the pair matrix is an $M\times N$ matrix consisting of the entailment scores $E_{ij}$. In Section~\ref{section:nli_granularity}, we examine the effect of granularity by splitting texts at the paragraph level or binning two sentences at a time. In Section~\ref{section:nli_category}, we explore the use of the contradiction and neutral categories in our experiments.

The example in Figure~\ref{fig:intro_example} has $M=4$ document sentences, and $N=3$ summary sentences, and the corresponding NLI Pair Matrix is the following:

$$
X_{pair} = \begin{bmatrix}
0.02 & 0.02 & 0.04\\
0.98 & 0.00 & 0.00\\
0.43 & 0.99 & 0.00\\
0.00 & 0.00 & 0.01\\
\end{bmatrix}
$$
The pair matrix can be interpreted as the weights of a bipartite graph which is also illustrated in Figure~\ref{fig:intro_example} where the opacity of each edge $(i,j)$ represents the entailment probability $E_{ij}$.

\begin{figure}
    \centering
    \includegraphics[width=0.47\textwidth]{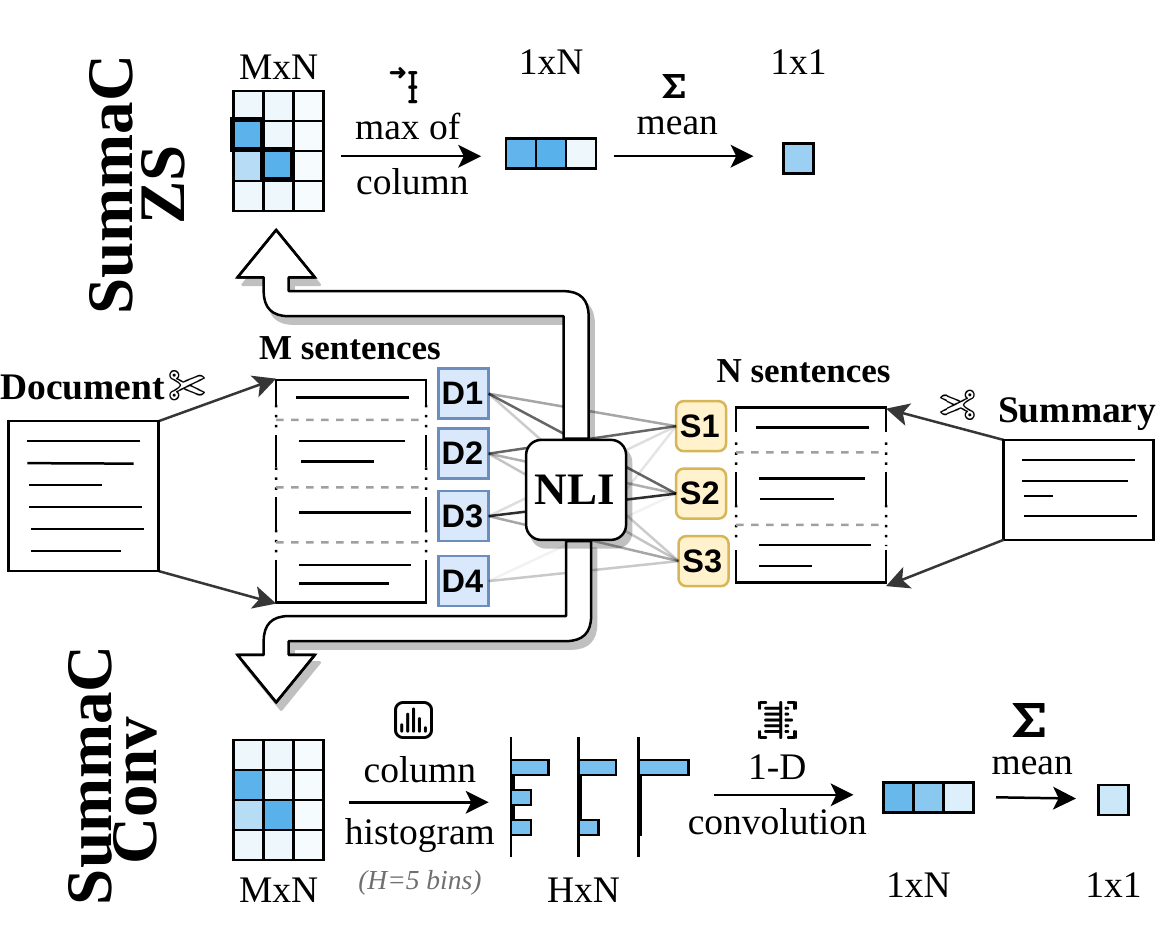}
    \caption{\textbf{Diagram of the \summaczs{} (top) and \summacconv{} (bottom) models.} Both models utilize the same NLI Pair Matrix (middle) but differ in their processing to obtain a score. The \summaczs{} is \textbf{Z}ero-\textbf{S}hot, and does not have trained parameters. \summacconv{} uses a convolutional layer trained on a binned version of the NLI Pair Matrix.}
    \label{fig:summac_models}
\end{figure}

The two \summac{} models take as input the same NLI Pair Matrix, but differ in the aggregation method to transform the pair matrix into a  score. Figure~\ref{fig:summac_models} presents an overview of \summaczs{} and \summacconv{}.

\subsection{\summaczs{}: Zero-Shot}

\begin{table*}[]
    \centering
    \resizebox{0.9\textwidth}{!}{%
    \begin{tabular}{lccccccc}
    \hline
    \textbf{Dataset} & \multicolumn{2}{c}{\textbf{Size}} & \textbf{\% Positive} & \textbf{IAA} & \textbf{Source} & \textbf{\# Summarizer} & \textbf{\# Sublabel} \\
     & \multicolumn{1}{l}{\textbf{Valid.}} & \multicolumn{1}{l}{\textbf{Test}} & \multicolumn{1}{l}{} & \multicolumn{1}{l}{} & \multicolumn{1}{l}{} & \multicolumn{1}{l}{} & \multicolumn{1}{l}{} \\ \hline
    CoGenSumm \cite{falke2019ranking} & 1281 & 400 & 49.8 & 0.65 & C & 3 & 0 \\
    XSumFaith \cite{maynez2020faithfulness} & 1250 & 1250 & 10.2 & 0.80 & X & 5 & 2 \\
    Polytope \cite{huang2020have} & 634 & 634 & 6.6 & - & C & 10 & 8 \\
    FactCC \cite{kryscinski2020evaluating} & 931 & 503 & 85.0 & - & C & 10 & 0 \\
    SummEval \cite{fabbri2021summeval} & 850 & 850 & 90.6 & 0.7 & C & 23 & 4 \\
    FRANK \cite{Pagnoni2021UnderstandingFI} & 671 & 1575 & 33.2 & 0.53 & C+X & 9 & 7 \\ \hline
    \end{tabular}
    }
    \caption{\textbf{Statistics of the six datasets in the \summac{} Benchmark.} For each dataset, we report the validation and test set sizes, the percentage of summaries with positive (consistent) labels (\textbf{\% Positive}), the inter-annotator agreement (when available, \textbf{IAA}), the source of the documents (\textbf{Source}: C for CNN/DM, X for XSum), the number of summarizers evaluated, and the number of sublabels annotated.}
    \label{table:summac_benchmark}
\end{table*}

In the \summaczs{} model, we reduce the pair matrix to a one-dimensional vector by taking the maximum (\texttt{max}) value of each column. On an intuitive level, for each summary sentence, this step consists of retaining the score for the document sentence that provides the strongest support for each summary sentence. For the example in Figure~\ref{fig:intro_example}:
$$
\mathrm{max}(X_{pair}, \textrm{axis=`col'}) = \begin{bmatrix}
0.98 & 0.99 & 0.04\\
\end{bmatrix}
$$

The second step consists of taking the \texttt{mean} of the produced vector, reducing the vector to a scalar which is used as the final model score. At a high level, this step aggregates sentence-level information into a single score for the entire summary. For example, in Figure~\ref{fig:intro_example}, the score produced by \summaczs{} would be $0.67$. If we removed the third sentence from the summary, the score would increase to $0.985$. We experiment with replacing the \texttt{max} and \texttt{mean} operators with other operators in Appendix~\ref{appendix:zs_operators}.

\subsection{\summacconv{}: Convolution}

One limitation of \summaczs{} is that it is highly sensitive to extrema, which can be noisy due to the presence of outliers and the imperfect nature of NLI models.  In \summacconv{}, we reduce the reliance on extrema values by instead taking into account the entire distribution of entailment scores for each summary sentence. For each summary sentence, a learned convolutional layer is in charge of converting the entire distribution into a single score.

The first step of the \summacconv{} algorithm is to turn each column of the NLI Pair Matrix into a fixed-size histogram that represents the distribution of scores for that given summary sentence.

We bin the NLI scores into $H$ evenly spaced bins (e.g., if $H=5$, the bins are $[0,0.2), [0.2,0.4), [0.4, 0.6), [0.6, 0.8), [0.8, 1)$).  Thus the first summary sentence of the example in Figure~\ref{fig:intro_example} would have the following histogram: $[2, 0, 1, 0, 1]$, because there are two values between $[0.0, 0.2]$ in the first column, one in $[0.4, 0.6]$ and one in $[0.8, 1.0]$.

By producing one histogram for each summary sentence, the binning process in the example of Figure~\ref{fig:intro_example} would produce:

$$
\mathrm{bin}(X_{pair}) = \begin{bmatrix}
2 & 3 & 4 \\
0 & 0 & 0 \\
1 & 0 & 0 \\
0 & 0 & 0 \\
1 & 1 & 0 \\
\end{bmatrix}
$$

The binned matrix is then passed through a 1-D convolution layer with a kernel size of $H$. The convolution layer scans the summary histograms one at a time, and compiles each into a scalar value for each summary. Finally, the scores of each summary sentence are averaged to obtain the final summary-level score.

In order to learn the weights of the convolution layer, we train the \summacconv{} model end-to-end with the synthetic training data in FactCC \cite{kryscinski2020evaluating}. The original training dataset contains one million \texttt{(document, summary)} pairs evenly distributed with consistent and inconsistent summaries. Because we are only training a small set of $H$ parameters (we use $H=50$), we find that using a 10,000 sub-sample is sufficient. We train the model using a cross-entropy loss, the Adam optimizer, a batch size of 32 and a learning rate of $10^{-2}$. We perform hyper-parameter tuning on a validation set from the FactCC dataset.

The number of bins used in the binning process, which corresponds to the number of parameters in the convolution layer, is also a hyper-parameter we tune on the validation set. We find that performance increases until 50 bins (i.e., a bin width of 0.02) and then plateaus. We use 50 bins in all our experiments.


\section{\summac{} Benchmark}
\label{section:summac_benchmark}

To rigorously evaluate the \summac{} models on a diverse set of summaries with consistency judgements, we introduce a new large benchmark dataset, the \summac{} Benchmark. It comprises the six largest available datasets for summary inconsistency detection, which we standardize to use the same classification task.

\subsection{Benchmark Standardization}

We standardize the task of summary inconsistency detection by casting it as a binary classification task. Each dataset contains \texttt{(document, summary, label)} samples, where the label can either be \textit{consistent} or \textbf{inconsistent}.

Each dataset is divided into a validation and test split, with the validation being available for parameter tuning. We used existing validation/test splits created by dataset authors when available. We did not find a split for XSumFaith, Polytope, and SummEval,  and created one by putting even-indexed samples in a validation split, and odd-indexed samples in the test split. This method of splitting maintains similar class imbalance and summarizer identity with the entire dataset.

We computed inter-annotator agreement calculated with Fleiss' Kappa \cite{fleiss1971measuring} on the dataset as an estimate for dataset quality, omitting datasets for which summaries only had a single annotator (Polytope and FactCC). Table~\ref{table:summac_benchmark} summarizes dataset statistics and properties.

\subsection{Benchmark Datasets}

We introduce each dataset in the benchmark chronologically, and describe the standardizing procedure.

\textbf{CoGenSumm} (\textbf{Co}rrectness of \textbf{Gen}erated \textbf{Summ}aries, CGS) \cite{falke2019ranking} is the first introduced dataset for summary inconsistency detection, based on models trained on the CNN/DM dataset \cite{nallapati2016abstractive}. The authors proposed that consistency detection should be approached as a ranking problem: given a consistent and inconsistent summary for a common document, a ranking model should score the consistent summary higher. Although innovative, other datasets in the benchmark do not always have positive and negative samples for a given document. We thus map the dataset to a classification task by using all inconsistent and consistent summaries as individual samples.

\textbf{XSumFaith} (e\textbf{X}treme \textbf{Sum}marization \textbf{Faith}fulness, XSF) \cite{maynez2020faithfulness} is a dataset with models trained on the XSum dataset \cite{narayan2018don} which consists of more abstractive summaries than CoGenSumm. The authors find that standard generators remain consistent for only 20-30\% of generated summaries. The authors differentiate between \textit{extrinsic} and \textit{intrinsic} hallucinations (which we call inconsistencies in this work). Extrinsic hallucinations, which involve words or concepts not in the original document can nonetheless be \textit{accurate} or \textit{inaccurate}. In order for a summarizer to generate an accurate extrinsic hallucination, the summarizer must possess external world knowledge. Because the authors found that the models are primarily inaccurate in terms of extrinsic hallucinations, we map both extrinsic and intrinsic hallucinations to a common inconsistent label.

\textbf{Polytope} \cite{huang2020have} introduces a more extensive typology of summarization errors, based on the Multi-dimensional Quality Metric \cite{mariana2014multidimensional}. Each summary is annotated with eight possible errors, as well as a severity level for the error. We standardize this dataset by labeling a summary as inconsistent if it was annotated with any of the five accuracy errors (and disregarded the three fluency errors). Each summary in Polytope was labeled by a single annotator, making it impossible to measure inter-annotator agreement.

\textbf{FactCC} \cite{kryscinski2020evaluating} contains validation and test splits that are entirely annotated by authors of the paper, because attempts at crowd-sourced annotation yielded low inter-annotator agreement. Prior work \cite{gillick2010non} shows that there can be divergence in annotations between experts and non-experts in summarization, and because the authors of the paper are NLP researchers familiar with the limitations of automatic summarizations, we expect that FactCC annotations differs in quality from other datasets. FactCC also introduces a synthetic dataset by modifying consistent summaries with semantically variant rules. We use a sub-portion of this synthetic dataset to train the \summacconv{} model.

\textbf{SummEval} \cite{fabbri2021summeval} contains summarizer outputs from seven extractive models and sixteen abstractive models. Each summary was labeled using a 5-point Likert scale along four categories: coherence, consistency, fluency, and relevance by 3 annotators. We label summaries as consistent if all annotators gave a score of 5 in consistency, and inconsistent otherwise. 

\textbf{FRANK} \cite{Pagnoni2021UnderstandingFI} contains annotations for summarizers trained on both CNN/DM and XSum, with each summary annotated by three crowd-workers. The authors propose a new typology with seven error types, organized into semantic frame errors, discourse errors and content verifiability errors. The authors confirm that models trained on the more abstractive XSum dataset generate a larger proportion of inconsistent summaries, compared to models trained on CNN/DM. We label summaries as consistent if a majority of annotators labeled the summary as containing no error.

\subsection{Benchmark Evaluation Metrics}

With each dataset in the \summac{} Benchmark converted to a binary classification task, we now discuss the choice of appropriate evaluation metrics for the benchmark. Previous work on each dataset in the benchmark used different evaluation methods, falling into three main categories.

First, CoGenSumm proposes a re-ranking based measure, requiring pairs of consistent and inconsistent summaries for any document evaluated; this information is not available in several datasets in the benchmark.

Second, XSumFaith, SummEval and FRANK report on correlation of various metrics with human annotations. Correlation has some advantages, such as not requiring a threshold and being compatible with the Likert-scale annotations of SummEval, however it is an uncommon choice to measure performance of a classifier due to the discrete and binary label.

Third, authors of FactCC measured model performance using binary F1 score, and balanced accuracy, which corrects unweighed accuracy with the class imbalance ratio, so that majority class voting obtains a score of 50\%.

The datasets have widely varying class imbalances, 
ranging from 6\% to 91\% positive samples. Therefore,
we select balanced accuracy \cite{brodersen2010balanced} as the primary evaluation metric for the \summac{} Benchmark. Balanced accuracy is defined as:
\begin{equation}
    B Acc = \frac12 \left( \frac{TP}{TP+FN} + \frac{TN}{TN + FP} \right)
\end{equation}
Where $TP$ stands for true positive, $FP$ false positive, $TN$ true negative, and $FN$ false negative. The choice of metric is based on the fact that accuracy is a conceptually simple, interpretable metric, and that adjusting the class imbalance out of the metric makes the score more uniform across datasets.

The balanced accuracy metric requires models to output a binary label (i.e., not a scalar score), which for most models requires the selection of a threshold in the score. The threshold is selected using the validation set, allowing for a different threshold for each dataset in the benchmark. Performance on the benchmark is the unweighted average of performance on the six datasets.

We choose Area Under the Curve of the Receiver Operating Chart (ROC-AUC) as a secondary evaluation metric, a common metric to summarize a classifier's performance at different threshold levels \cite{bradley1997use}.

\section{Results}
\label{section:results}

\begin{table*}[]
    \centering
    \resizebox{0.99\textwidth}{!}{%
    \begin{tabular}{
    llcccccccc}
    & & \multicolumn{6}{c}{\textbf{\summac{} Benchmark Datasets}} &  & \\
    \cmidrule(lr){3-8}
    \textbf{Model Type} & \textbf{Model Name} & \textbf{CGS} & \textbf{XSF} & \textbf{Polytope} & \textbf{FactCC} & \textbf{SummEval} & \textbf{FRANK} & \textbf{Overall} & \textbf{Doc./min.} \\ 
    \cmidrule(r){1-2} \cmidrule(lr){3-8} \cmidrule(lr){9-9} \cmidrule(l){10-10}
    
    \multirow{2}{*}{Baseline} & NER-Overlap & 53.0 & 63.3 & 52.0 & 55.0 & 56.8 & 60.9 & 56.8 & 55,900 \\
     & MNLI-doc & 57.6 & 57.5 & 61.0 & 61.3 & 66.6 & 63.6 & 61.3 & 6,200 \\
     \cmidrule(r){1-2} \cmidrule(lr){3-8} \cmidrule(lr){9-9} \cmidrule(l){10-10}
    Classifier & FactCC-CLS & 63.1 & 57.6 & 61.0 & 75.9 & 60.1 & 59.4 & 62.8 & 13,900 \\
    \cmidrule(r){1-2} \cmidrule(lr){3-8} \cmidrule(lr){9-9} \cmidrule(l){10-10}
    Parsing & DAE & 63.4 & 50.8 & 62.8 & 75.9 & 70.3 & 61.7 & 64.2 & 755 \\
    \cmidrule(r){1-2} \cmidrule(lr){3-8} \cmidrule(lr){9-9} \cmidrule(l){10-10}
    \multirow{2}{*}{QAG} & FEQA & 61.0 & 56.0 & 57.8 & 53.6 & 53.8 & 69.9 & 58.7 & 33.9 \\
     & QuestEval & 62.6 & 62.1 & \hspace{1ex}\textbf{70.3}* & 66.6 & 72.5 & \textbf{82.1} & 69.4 & 22.7 \\
    \cmidrule(r){1-2} \cmidrule(lr){3-8} \cmidrule(lr){9-9} \cmidrule(l){10-10}
    \multirow{2}{*}{NLI} & \summaczs{} & \hspace{1.0ex}\textbf{70.4}* & 58.4 & 62.0 & \hspace{1.0ex}83.8* & 78.7 & 79.0 & \hspace{1.0ex}72.1* & 435 \\
     & \summacconv{} & 64.7 & \hspace{1.0ex}\textbf{66.4}* & 62.7 & \hspace{2.0ex}\textbf{89.5}** & \hspace{2.0ex}\textbf{81.7}** & \textbf{81.6} & \hspace{2.0ex}\textbf{74.4}** & 433 \\
    \bottomrule
    \end{tabular}
    }
    \caption{\textbf{Performance of Summary Inconsistency Detection models on the test set of the \summac{} Benchmark.} Balanced accuracy is computed for each model on the six datasets in the benchmark, and the average is computed as the overall performance on the benchmark. We obtain confidence intervals comparing the \summac{} models to prior work: * indicates an improvement with 95\% confidence, and ** 99\% confidence (details in Section~\ref{section:stat_testing}). The results of the throughput analysis of Section~\ref{section:throughput_analysis} are in column Doc./min (Documents per minute).}
    \label{table:benchmark_results}
\end{table*}

We compared the \summac{} models against a  wide array of baselines and state-of-the-art methods.

\subsection{Comparison Models}

We evaluated the following models on the \summac{} Benchmark:

\textbf{NER Overlap} uses the spaCy named entity recognition (NER) model \cite{spacy} to detect when an entity present in the summary is not present in the document. This model, adapted from \cite{laban-etal-2021-keep}, considers only a subset of entity types as hallucinations (i.e., \textit{PERSON, LOCATION, ORGANIZATION}, etc.)

\textbf{MNLI-doc} is a Roberta \cite{liu2019roberta} model finetuned on the MNLI dataset \cite{williams2018broad}. The document is used as the premise and the summary as a hypothesis, and we use the predicted probability of entailment as a score, similar to prior work on using NLI models for inconsistency detection \cite{kryscinski2020evaluating}.

\textbf{FactCC-CLS} is a Roberta-base model finetuned on the synthetic training portion of the FactCC dataset. Although trained solely on artificially created inconsistent summaries, prior work showed the model to be competitive on the FactCC and FRANK datasets.

\textbf{DAE} \cite{goyal2020evaluating} is a parsing-based model using the default model and hyper-parameters provided by the authors of the paper\footnote{https://github.com/tagoyal/dae-factuality}.

\textbf{FEQA} \cite{durmus2020feqa} is a QAG method, using the default model and hyper-parameters provided by the authors of the paper\footnote{https://github.com/esdurmus/feqa}.

\textbf{QuestEval} \cite{scialom2021questeval} is a QAG method taking both precision and recall into account. We use the default model and hyper-parameters provided by the authors of the paper\footnote{https://github.com/ThomasScialom/QuestEval}. The model has an option to use an additional question weighter, however experiments revealed that the weighter lowered overall performance on the validation portion of the \summac{} Benchmark, and we compare to the model without weighter.

\subsection{\summac{} Benchmark Results}
\label{section:benchmark_results}

Balanced accuracy results are summarized in Table~\ref{table:benchmark_results}. We find that the \summac{} models achieve the two best performances in the benchmark. \summacconv{} achieves the best benchmark performance at 74.4\%, 5 points above QuestEval, the best method not involving NLI.

Looking at the models' ability to  generalize across datasets and varying scenarios of inconsistency detection provides interesting insights. For example, the FactCC-CLS model achieves strong performance on the FactCC dataset, but close to lowest performance on FRANK and XSumFaith. In comparison, \summac{} model performance is strong across the board.

The strong improvement from the \summaczs{} to \summacconv{} also shines a light on the importance of considering the entire distribution of document scores for each summary sentence, instead of taking only the maximum score: the \summacconv{} model learns to look at the distribution and makes more robust decisions, leading to gains in performance.

The table of results with the ROC-AUC metric, the secondary metric of the \summac{} Benchmark is included in Appendix~\ref{table:benchmark_results_roc}, echoing the trends seen with the balanced accuracy metric.

\subsubsection{Statistical Testing}
\label{section:stat_testing}

We aim to determine whether the performance improvements of the \summac{} models are statistically significant. For each dataset of the benchmark, we perform two tests through bootstrap resampling \cite{efron1982jackknife}, comparing each of the \summac{} models to the best-performing model from prior work. We perform interval comparison at two significance level: $p=0.05$ and $p=0.01$, and apply the Bonferroni correction \cite{bonferroni1935calcolo} as we perform several tests on each dataset. We summarize which improvements are significant in Table~\ref{table:benchmark_results}, and perform a similar testing procedure for the ROC-AUC results in Table~\ref{table:benchmark_results_roc}.

\summac{} models lead to a statistically significant improvement on CoGenSumm, XSumFaith, FactCC and SummEval. QuestEval outperforms the \summac{} models on Polytope at a confidence of 95\%. On the FRANK dataset, QuestEval and \summacconv{} achieve highest performance with no statistical difference. Overall on the benchmark, both \summac{} models significantly outperform prior work, \summaczs{} at a $p=0.05$ significance level and \summacconv{} at $p=0.01$.

\subsubsection{Computational Cost Comparison}
\label{section:throughput_analysis}
Computational cost of the method is an important practical factor to consider when choosing a model to use, as some applications such as training with a generator with Reinforcement Learning might require a minimum throughput from the model (i.e., number of documents processed by the model per unit of time).

A common method to compare algorithms is using computational complexity analysis, computing the amount of resources (time, space) needed as the size of the input varies. Computational complexity analysis is impractical in our case, as the units of analysis differ between models, and do not allow for a direct comparison. More specifically, some of the model's complexity scales with the number of sub-word units in the document (\texttt{MNLI-doc}, \texttt{FactCC-CLS}), some with the number of entities in a document (\texttt{NER-Overlap}, \texttt{DAE}, \texttt{QuestEval}), and some with number of sentences (the \summac{} models).

We instead compare models by measuring throughput on a fixed dataset using a common hardware setup. More precisely, we measured the processing time of each model on the 503 documents in the test set of FactCC (with an average of 33.2 sentences per document), running a single Quadro RTX 8000 GPU. For prior work, we used implementation publicly released by the authors, and made a best effort to use the model at an appropriate batch size for a fair comparison.

The result of the throughput analysis is included in Table~\ref{table:benchmark_results} (column Docs./min.). \summac{} models are able to process around 430 documents per minute, which is much lower than some of the baselines capable of processing more than 10,000 documents per minute. However, QAG methods are more than 10 times slower than \summac{} models, processing only 20-40 documents per minute.

\subsection{Further Results}
We now examine how different components and design choices affect \summac{} model performance.

\subsubsection{Choice of NLI Model}

\begin{table}[]
    \resizebox{0.47\textwidth}{!}{%
    \begin{tabular}{lccc}
     &  & \multicolumn{2}{c}{\textbf{Performance}} \\
     \cmidrule(l){3-4}
    \textbf{Architecture} & \textbf{NLI Dataset} & \textbf{ZS} & \textbf{Conv} \\
    \cmidrule(r){1-2} \cmidrule(l){3-4}
    Dec. Attn & SNLI & 56.9 & 56.4 \\
    \cmidrule(r){1-2} \cmidrule(l){3-4}
     & SNLI & 66.6 & 64.0 \\
    BERT Base & MNLI & 69.5 & 69.8 \\
     & MNLI+VitaminC & 67.9 & 71.2 \\
    \cmidrule(r){1-2} \cmidrule(l){3-4}
     & SNLI & 66.6 & 62.4 \\
     & SNLI+MNLI+ANLI & 69.9 & 71.7 \\
    BERT Large & VitaminC & 71.1 & 72.8 \\
     & MNLI & 70.9 & 73.0 \\
     & MNLI+VitaminC & \textbf{72.1} & \textbf{74.4} \\
    \bottomrule
    \end{tabular}
    }
    \caption{\textbf{Effect of NLI model choice on \summac{} models performance.} For each NLI model, we include the balanced accuracy scores of \summaczs{} and \summacconv{}. BERT X corresponds to a BERT or other pre-trained models of similar size.}
    \label{table:nli_choice}
\end{table}

\summac{} models rely on an NLI model at their core, which consists of choosing two main components: a model architecture, and a dataset to train on. We investigate the effect of both of these choices on the performance of \summac{} models on the benchmark.

Regarding model architectures, we experiment with the decomposable attention model \cite{parikh2016decomposable}, which is a pre-Transformer architecture model that was shown to achieve high performance on SNLI, as well as Transformer base and Transformer Large architectures.

With respect to datasets, we include models trained on standard NLI datasets such as  SNLI \cite{bowman2015large} and MNLI \cite{williams2018broad}, as well as more recent datasets such as Adversarial NLI \cite{nie2019adversarial} and Vitamin C \cite{schuster2021get}.

Results are summarized in Table~\ref{table:nli_choice}, and we emphasize three trends. 
First, the low performance of the decomposable attention model used in experiments in prior work \cite{falke2019ranking}, confirms that less recent NLI models did not transfer well to summary inconsistency detection.

Second, NLI models based on pre-trained Transformer architectures all achieve strong performance on the benchmark, and average increase of 1.3 percentage points when going from a base to a large architecture.

Third, the choice of NLI dataset has a strong influence on overall performance. SNLI leads to lowest performance, which is expected as its textual domain is based on image captions, which are dissimilar to the news domain. MNLI and Vitamin C trained models both achieve close to the best performance, and training on both jointly leads to the best model, which we designate as the default NLI model for the \summac{} models (i.e., the model included in Table~\ref{table:benchmark_results}).

The latter two trends point to the fact that improvements in the field of NLI lead to improvements in the \summac{} models, and we can expect that future progress in the NLI community will translate to gains of performance when integrated into the \summac{} model.

We  relied on trained models available in HuggingFace's Model Hub \cite{wolf-etal-2020-transformers}. Details in Appendix~\ref{appendix:hf_model_card}.

\subsubsection{Choice of NLI Category}
\label{section:nli_category}

\begin{table}[]
\centering
    \resizebox{0.42\textwidth}{!}{%
    \begin{tabular}{cccccc}
    \multicolumn{3}{c}{\textbf{Category}} & \multicolumn{3}{c}{\textbf{\summacconv{} Performance}} \\
    \cmidrule(r){1-3} \cmidrule(l){4-6}
    \textbf{E} & \textbf{N} & \textbf{C} & \textbf{VITC+MNLI} & \textbf{ANLI} & \textbf{MNLI} \\ 
    \cmidrule(r){1-3} \cmidrule(l){4-6}
    \ding{51} & & & \textbf{74.4} & 69.2 & 72.6 \\
     & \ding{51} & & 71.2 & 55.8 & 66.4 \\
     & & \ding{51} & 72.5 & 69.2 & 72.6 \\
    \ding{51} & \ding{51} & & 73.1 & 69.6 & 72.6 \\
    \ding{51} & & \ding{51} & 74.0 & \textbf{70.2} & \textbf{73.0} \\
     & \ding{51} & \ding{51} & 72.5 & 69.2 & 72.6 \\
    \ding{51} & \ding{51} & \ding{51} & 74.0 & 69.7 & \textbf{73.0} \\
    \bottomrule
    \end{tabular}
    }
    \caption{\textbf{Effect of NLI category inclusion on \summacconv{} performance.} Models had access to different subsets of the three category predictions (\textbf{E}ntailment, \textbf{N}eutral, \textbf{C}ontradiction), with performance measured in terms of balanced accuracy. Experiments were performed with 3 NLI models: Vitamic C+MNLI, ANLI and MNLI.}
    \label{table:nli_labels}
\end{table}

The NLI task is a three-way classification task, yet most prior work has limited usage of the model to the use of the entailment probability for inconsistency detection \cite{kryscinski2020evaluating, falke2019ranking}. We run a systematic experiment by training multiple \summacconv{} models which have access to varying subsets of the NLI labels, and measure the impact on overall performance. Results are summarized in Table~\ref{table:nli_labels}. Using solely the entailment category leads to strong performance for all models. However, explicitly including the contradiction label as well leads to small boosts in performance for the ANLI and MNLI models. 

With future NLI models being potentially more nuanced and calibrated, it is possible that inconsistency detectors models will be able to rely on scores from several categories.

\subsubsection{Choice of Granularity}
\label{section:nli_granularity}

\begin{table}
    \resizebox{0.47\textwidth}{!}{%
    \begin{tabular}{cccccc}
     & & \multicolumn{4}{c}{\textbf{Performance}} \\
    \cmidrule(l){3-6}
    \multicolumn{2}{c}{\textbf{Granularity}} & \multicolumn{2}{c}{\textbf{MNLI}} & \multicolumn{2}{c}{\textbf{MNLI + VitC}} \\
    \cmidrule(r){1-2} \cmidrule(lr){3-4} \cmidrule(l){5-6}
    \textbf{Document} & \textbf{Summary} & \textbf{ZS} & \textbf{Conv} & \textbf{ZS} & \textbf{Conv} \\ 
    \cmidrule(r){1-2} \cmidrule(lr){3-4} \cmidrule(l){5-6}
    \multirow{2}{*}{Full} & Full  & 56.4 & - & 72.1 & - \\
     & Sentence  & 57.4 & - & 73.1 & - \\
    \cmidrule(r){1-2} \cmidrule(lr){3-4} \cmidrule(l){5-6}
    \multirow{2}{*}{Paragraph} & Full & 59.8 & 61.8 & 69.8 & 71.2 \\
     & Sentence & 65.2 & 64.7 & 72.6 & 74.3 \\
    \cmidrule(r){1-2} \cmidrule(lr){3-4} \cmidrule(l){5-6}
    \multirow{2}{*}{Two Sent.} & Full & 64.0 & 63.8 & 69.7 & 71.3 \\
     & Sentence & 71.2 & \textbf{73.5} & 72.5 & \textbf{74.7} \\
    \cmidrule(r){1-2} \cmidrule(lr){3-4} \cmidrule(l){5-6}
    \multirow{2}{*}{Sentence} & Full & 58.7 & 61.1 & 68.4 & 69.4 \\
     & Sentence  & 70.3 & \textbf{73.0} & 72.1 & \textbf{74.4} \\
    \bottomrule
    \end{tabular}
    }
    \caption{\textbf{Effect of granularity choice on \summac{} models performance.} We tested four granularities on the document side: full, paragraph, two sentence and sentence, and two granularities on the summary side: full and sentence. Performance of the four models is measured in balanced accuracy on the benchmark test set.}
    \label{table:granularity_analysis}
\end{table}

So far, we've reported experiments primarily with a sentence-level granularity, as it matches the granularity of NLI datasets. One can imagine cases where sentence-level granularity might be limiting. For example, in the case of a summary performing a \textit{sentence fusion} operation, an NLI model might not be able to correctly predict entailment of the fused sentence, seeing only one sentence at a time.

To explore this facet further, we experiment with modifying the granularity of both the document and the summary. With regards to document granularity, we consider four granularities: (1) \textbf{full text}, the text is treated as a single block, (2) \textbf{paragraph-level} granularity, the text is separated into paragraph blocks, (3) \textbf{two-sentence} granularity, the text is separated into blocks of contiguous sentences of size two (i.e., block 1 contains sentence 1-2, block 2 contains sentence 3-4), and (4) \textbf{sentence-level}, splitting text at individual sentences. For the summary granularity, we only consider two granularities: (1) \textbf{full text}, and (2) \textbf{sentence}, because other granularities are less applicable since summaries usually consist of three sentences or fewer. 

We study the total of 8 \texttt{(document, summary)} granularity combinations with the two best-performing NLI models of Table~\ref{table:benchmark_results}: MNLI and Vitamin C, each included as \summaczs{} and \summacconv{} models.\footnote{We skip \summacconv{} experiments involving full text granularity on the document-side, as that case reduces the binning process to having a single non-zero value.}

Results for the granularity experiments are summarized in Table~\ref{table:granularity_analysis}. Overall, finer granularities lead to better performance, with \texttt{(sentence,sentence)} and \texttt{(two sent, sentence)} achieving highest performance across all four models.

The MNLI-only trained model achieves lowest performance when used with full text granularity on the document level, and performance steadily increases from 56.4\% to 73.5\% as granularity is made finer both on the document and summary side. Results for the MNLI+VitaminC model vary less with changing granularity, showcasing that the model is perhaps more robust to different granularity levels. However the \texttt{(two sent, sentence)} and \texttt{(sentence,sentence)} settings achieve highest performance, implying that finer granularity remains valuable.

For all models, performance degrades in cases where granularity on the document level is finer than summary granularity. For example the \texttt{(sentence, full)} or \texttt{(two sent, full)} combinations lead to some of the lowest performance. This is expected, as in cases in which summaries have several sentences, it is unlikely that they will fully be entailed by a single document sentence. This implies that granularity on the document side should be coarser or equal the summary's granularity.

Overall, we find that finer granularity for the document and summary is beneficial in terms of performance and recommend the use of a \texttt{(sentence, sentence)} granularity combination.

\section{Discussion and Future Work}

\textbf{Improvements on the Benchmark.} The models we introduced in this paper are just a first step towards harnessing NLI models for inconsistency detection. Future work could explore a number of improvements: combining the predictions of multiple NLI models, or combining multiple granularitiy levels, for example through multi-hop reasoning \cite{zhao2019transformer}.

\textbf{Interpretability of model output}. If a model can pinpoint which portion of a summary is inconsistent, some work has shown that corrector models can effectively re-write the problematic portions and often remove the inconsistency \cite{dong2020multi}. Furthermore, fine-grained consistency scores can be incorporated into visual analysis tools for summarization such as SummViz \cite{vig2021summvis}. The \summaczs{} model is directly interpretable, whereas the \summacconv{} is slightly more opaque, due to the inability to trace back a low score to a single sentence in the document being invalidated. Improving the interpretability of the \summacconv{} model is another open area for future work.

\textbf{Beyond news summarization}. The six datasets in the \summac{} Benchmark contain summaries from the news domain, one of the most common application of summarization technology. Recent efforts to expand the application of summarization to new domains such as legal \cite{kornilova2019billsum} or scholarly \cite{cachola2020tldr} text will hopefully lead to the study of inconsistency detection in these novel domains, and perhaps even out of summarization on tasks such as text simplification, or code generation.

\textbf{Towards Consistent Summarization}. Inconsistency detection is but a first step in eliminating inconsistencies from summarization. Future work can include more powerful inconsistency detectors in the training of next generation summarizers to reduce the prevalence of inconsistencies in generated text.

\section{Conclusion}

We introduce \summaczs{} and \summacconv{}, two NLI-based models for summary inconsistency detection based on the key insight that NLI models require sentence-level input to work best. Both models achieve strong performance on the \summac{} Benchmark, a new diverse and standardized collection of the six largest datasets for inconsistency detection. \summacconv{} outperforms all prior work with a balanced accuracy score of 74.4\%, an improvement of five absolute percentage points over the best baseline. To the best of our knowledge, this the first successful attempt at adapting NLI models for inconsistency detection, and we believe that there are many exciting opportunities for further improvements and applications of our methods.

\section*{Acknowledgments}

We would like to thank Katie Stasaski, Dongyeop Kang, the TACL reviewers and editors for their helpful comments, as well as Artidoro Pagnoni for helpful pointers during the project. This work was supported by a Microsoft BAIR Commons grant as well as a Microsoft Azure Sponsorship.


\bibliography{tacl2018}
\bibliographystyle{acl_natbib}

\clearpage

\appendix
\section*{Appendix}
\renewcommand{\thetable}{A\arabic{table}}
\setcounter{table}{0}
\renewcommand{\thefigure}{A\arabic{figure}}
\setcounter{figure}{0}

\section{NLI Model Origin}
\label{appendix:hf_model_card}

We list the NLI models we used throughout the paper, which can be retrieved on HuggingFace's model hub\footnote{https://huggingface.co/models}. BERT stands for any Pre-trained bi-directional Transformer of an equivalent size:
\begin{itemize}
    \setlength\itemsep{-0.3em}
    \item \texttt{boychaboy/SNLI\_roberta-base} BERT Base+SNLI
    \item \texttt{microsoft/deberta-base-mnli} BERT Base+MNLI
    \item \texttt{tals/albert-base-vitaminc-mnli} BERT Base + MNLI + VitaminC
    \item \texttt{boychaboy/SNLI\_roberta-large} BERT Large+SNLI
    \item \texttt{tals/albert-xlarge-vitaminc} Bert Large+VitaminC
    \item \texttt{roberta-large-mnli} \hspace{0.2\textwidth} Bert Large+MNLI
    \item \texttt{tals/albert-xlarge-vitaminc-mnli} BERT Large+MNLI+VitaminC
    
\end{itemize}

\section{\summaczs{} Operator Choice}
\label{appendix:zs_operators}

\begin{table}
    \centering
    \resizebox{0.33\textwidth}{!}{%
    \begin{tabular}{cccc}
    & \multicolumn{3}{c}{\textbf{Operator 2}}     \\
    \cmidrule(l){2-4}
    \textbf{Op. 1} & \textbf{Min} & \textbf{Mean} & \textbf{Max} \\
    \cmidrule(r){1-1} \cmidrule(l){2-4}
    \textbf{Min} & 53.1 & 55.7 & 57.4 \\
    \textbf{Mean} & 60.5 & 62.8 & 62.0 \\
    \textbf{Max} & 68.8 & \textbf{72.1} & 69.1 \\
    \bottomrule
    \end{tabular}
    }
    \caption{\textbf{Effect of operator choice on the performance of the \summaczs{} model, measured in terms of balanced accuracy.} \texttt{Operator 1} reduces the row dimension of the NLI Pair Matrix, and \texttt{Operator 2} reduces the column dimension.}
    \label{table:appendix_zs_operator}
\end{table}

Table~\ref{table:appendix_zs_operator} measures the effect of the choice of the two operators in the \summaczs{} model. We explore three options (\texttt{min, mean} and \texttt{max}) for each operator. We find that the choice of \texttt{max} for Operator 1 and \texttt{mean} for Operator 2 achieves the highest performance and use these choices in our model.

\section{\summac{} Benchmark ROC-AUC Results}

Table~\ref{table:benchmark_results_roc} details results of models on the benchmark according to the ROC-AUC metric, confirming that the \summac{} models achieve the two best accuracy results on the benchmark.


\begin{table*}[!htbp]
    \centering
    \resizebox{0.98\textwidth}{!}{%
    \begin{tabular}{llccccccc}
        & & \multicolumn{6}{c}{\textbf{\summac{} Benchmark Datasets}} &  \\
    \cmidrule(lr){3-8}
    \textbf{Model Type} & \textbf{Model Name} & \textbf{CGS} & \textbf{XSF} & \textbf{Polytope} & \textbf{FactCC} & \textbf{SummEval} & \textbf{FRANK} & \textbf{Overall} \\
    \cmidrule(r){1-2} \cmidrule(lr){3-8} \cmidrule(l){9-9}
    \multirow{2}{*}{Baseline} & NER-Overlap & 53.0 & 61.7 & 51.6 & 53.1 & 56.8 & 60.9 & 56.2 \\
     & MNLI-doc & 59.4 & 59.4 & 62.6 & 62.1 & 70.0 & 67.2 & 63.4 \\
    \cmidrule(r){1-2} \cmidrule(lr){3-8} \cmidrule(l){9-9}
    Classifier & FactCC-CLS & 65.0 & 59.2 & 63.5 & 79.6 & 61.4 & 62.7 & 65.2 \\
    \cmidrule(r){1-2} \cmidrule(lr){3-8} \cmidrule(l){9-9}
    Parsing & DAE & 67.8 & 41.3 & 64.1 & 82.7 & 77.4 & 64.3 & 66.3 \\
    \cmidrule(r){1-2} \cmidrule(lr){3-8} \cmidrule(l){9-9}
    \multirow{2}{*}{QAG} & FEQA & 60.8 & 53.4 & 54.6 & 50.7 & 52.2 & 74.8 & 57.7 \\
     & QuestEval & 64.4 & 66.4 & \textbf{72.2} & 71.5 & 79.0 & \textbf{87.9} & 73.6 \\
    \cmidrule(r){1-2} \cmidrule(lr){3-8} \cmidrule(l){9-9}
    \multirow{2}{*}{NLI} & \summaczs{} & \textbf{73.1} & 58.0 & 60.3 & 83.7 & 85.5 & 85.3 & 74.3 \\
     & \summacconv{} & 67.6 & \textbf{70.2} & 62.4 & \hspace{2.0ex}\textbf{92.2}** & \hspace{1.0ex}\textbf{86.0*} & \textbf{88.4} & \hspace{2.0ex}\textbf{77.8}** \\
     \bottomrule
    \end{tabular}
    }
    \caption{\textbf{Performance of Summary Inconsistency Detection models on the test portion of the \summac{} Benchmark in terms of ROC-AUC metric.} The metric is computed for each model on the six datasets in the benchmark, and the average is computed as the overall performance on the benchmark. Confidence intervals comparing the \summac{} models to prior work: * indicates an improvement with 95\% confidence, and ** 99\% confidence (details in Section~\ref{section:stat_testing})}
    \label{table:benchmark_results_roc}
\end{table*}


\end{document}